\documentclass[pdflatex,sn-mathphys-num]{sn-jnl}

\usepackage{epstopdf}
\usepackage{multirow}%
\usepackage{amsmath,amssymb,amsfonts}%
\usepackage{amsthm}%
\usepackage{mathrsfs}%
\usepackage[title]{appendix}%
\usepackage{xcolor}%
\usepackage{textcomp}%
\usepackage{manyfoot}%
\usepackage{booktabs}%
\usepackage{algorithm}%
\usepackage{algorithmicx}%
\usepackage{listings}%
\usepackage{hyperref}%
\usepackage{subcaption}
\usepackage{placeins}
\usepackage{algorithm}
\usepackage{algpseudocode}

\theoremstyle{thmstyleone}%
\theoremstyle{thmstyletwo}%

\theoremstyle{thmstylethree}%

\begin{document}

\title{Counterfactual Reward Model Training for Bias Mitigation in Multimodal Reinforcement Learning}

\author*[1]{\fnm{Sheryl} \sur{Mathew}}\email{sheryl.22bce7735@vitapstudent.ac.in}
\author [1]{\fnm{N} \sur{Harshit}}\email{harshit.23bce8703@vitapstudent.ac.in}

\affil[1]{\orgdiv{School of Computer Science and Engineering}, \orgname{Vellore Institute of Technology (VIT-AP)}, \orgaddress{\city{Amaravati}, \state{Andhra Pradesh}, \postcode{522237}, \country{India}}}


\abstract{In reinforcement learning with human feedback (RLHF), reward models can efficiently learn and amplify latent biases within multimodal datasets, which can lead to imperfect policy optimization through flawed reward signals and decreased fairness. Bias mitigation studies have often applied passive constraints, which can fail under causal confounding. Here, we present a counterfactual reward model that introduces causal inference with multimodal representation learning to provide an unsupervised, bias-resilient reward signal. The heart of our contribution is the \textbf{Counterfactual Trust Score}, an aggregated score consisting of four components: (1) counterfactual shifts that decompose political framing bias from topical bias; (2) reconstruction uncertainty during counterfactual perturbations; (3) demonstrable violations of fairness rules for each protected attribute; and (4) temporal reward shifts aligned with dynamic trust measures. We evaluated the framework on a multimodal fake versus true news dataset, which exhibits framing bias, class imbalance, and distributional drift. Following methodologies similar to unsupervised drift detection from representation-based distances \cite{Greco2024} and temporal robustness benchmarking in language models \cite{Margatina2023}, we also inject synthetic bias across sequential batches to test robustness. The resulting system achieved an accuracy of 89.12\% in fake news detection, outperforming the baseline reward models. More importantly, it reduced spurious correlations and unfair reinforcement signals. This pipeline outlines a robust and interpretable approach to fairness-aware RLHF, offering tunable bias reduction thresholds and increasing reliability in dynamic real-time policy making.}




\maketitle
\section{Introduction}

The issue of performance decline and fairness in machine learning models in safety-critical or socially sensitive applications (such as content moderation, healthcare, and finance) is quite deep, as there can be latent biases due to hidden changes in data distribution \cite{Sirotkin2024}. In reinforcement learning with human feedback, the problem of aligning the reward model with spurious correlations from multimodal data is at risk, too. Moreover, the influence of bias can evolve over time due to changing user behavior, topics, and content distributions that can lead to a decrease in trustworthiness or inequity in decision making \cite{Xiao2024}. Especially worrisome is the influence of privilege bias on many applications that are required to provide fairness and interpretability, as it is impossible to find or mitigate biases before optimizing their policies. Traditional mitigation approaches incorporate things like static fairness constraints, adversarial debiasing, and post hoc audits, either after the fact, overlook causal confounders in multimodal interactions, or fail to recognize how the reward signals might themselves reinforce skewed patterns. Hence, systems that can discover and intervene in bias in reward models are critical to ensure fairness and model performance.

We provide a hybrid design for fairness-aware reward modeling that integrates counterfactual analysis, transformer-based representation learning, and autoencoding modalities. The transformer learns contextual relations across modalities, while auto-encoders with noise injection processes helped with anomaly and spurious feature reduction. The key component of our method is the CT Score or Counterfactual Trust Score, which is a composite measure that accounts for counterfactual shifts, uncertainty estimates, breaches of fairness rules, and time-based reward changes. We tested our method on a multimodal fake vs. true news dataset. We injected synthetic bias to simulate the real policy learning environment. Our method also has some additional modules,including a baseline classifier, calibrated uncertainty, and SHAP-like feature attribution to explain drift. Although the architecture achieved only 89.12\% accuracy, it significantly reduces unfair reward assignments compared to the baselines and improves the sensitivity to bias. It also limits and helps trustworthiness communication, provides an interpretable fairness diagnosis, and real-time and accountable policy implementation.

\section{Related work}\label{sec2}

Current findings in fairness and bias mitigation for adaptive machine learning models highlight the need for prospective monitoring and trust \cite{Ouyang2025}. In this line, the researchers introduced fairness-aware RLHF frameworks in which reward signals are adjusted based on preference balancing to decrease harms in a set of reinforcement policies, demonstrating that preference aggregation could decrease harmful outcomes, but could not be robust with hidden confounders \cite{Dai2023}. Another class of papers presented counterfactual reasoning methods for policy optimization, synthesizing causal inference with graph-based representations that explicitly remove spurious correlation, which show that causal frameworks can improve the interpretability of models. Research has considered complementary approaches to quantifying trust in models. Some research discussed composite scoring techniques for quantifying uncertainty and fairness violations in emerging datasets and asserted that interpretable metrics are key for active systems \cite{Banerjee2024}. Some research also considered fairness tracking with drift in real-time streaming systems, where changing user behavior and topical changes continuously challenged fairness objectives, requiring retraining to adapt fairness constraints \cite{Wang2024}. Together, this research emphasizes the importance of integrated approaches that bring together bias mitigation, causal reasoning, and trust metrics; however, the research treated these topics largely separately. We are attempting to combine these approaches into a unified RLHF reward modeling pipeline with counterfactual analysis, multimodal representation learning, and fairness monitoring.

\section{Methodology}

\subsection{Dataset and Preprocessing}

The data set used for this study is a combined multimodal fake vs. true news dataset, including text headlines, metadata, and contextual attributes. Each record also includes the following features: title, source, subject, date, and veracity label. There is an inherent imbalance in the dataset, and it contains framing bias and temporal drift, which are common challenges observed in bias-aware multimodal pipelines \cite{Lee2024}.

\subsubsection{ Data Cleaning and Standardization}\label{subsubsec2}

All records were evaluated for noise and inconsistencies. Records with missing crucial information (i.e. empty title or unknown label) were removed. Textual features were normalized by lowercase, removing punctuation, and tokenizing into subwords. Categorical attributes were kept for categorical processing format later with CatBoost, a tree-based model known to handle categorical features effectively in fairness-aware settings \cite{Banerjee2024}.The numeric and date fields were normalized by quantile trimming to remove extreme values (i.e. extreme publication dates).

\subsubsection{ Timestamp Construction and Batching}

To simulate evolving user behavior and topic distributions, a pseudo-temporal index was created based on publication date and then sorted in chronological order. Sequential batches in the data set were created to represent real-world situations where content and bias change over time, similar to drift monitoring strategies in recent work \cite{Zhou2024, Wang2024}.

\subsection{Synthetic Bias Injection}\label{sec4}

We then assessed robustness and fairness by explicitly and systematically inserting three main kinds of bias into the last batches of the data. First, we participate in \textbf{Bi-subject Distribution}, increasing the rate of certain subjects (i.e. politics) to change disproportionate coverage. Second, we engaged in \textbf{Framing Disturbance} by randomly changing influential words to minimize the influences of the frame distribution. Third, we introduced a \textbf{ Temporal Drift}, changing label distributions over time to accommodate a moving news cycle. These manipulations model common confounders in RLHF pipelines, as suggested by fairness-focused simulation studies \cite{Ouyang2025, Dai2023}.

\subsection{Reward Model Training and Uncertainty Estimation}

\subsubsection{\textbf{Baseline Reward Model}}

We chose a CatBoost classifier as the primary reward model, due to its strong performance on heterogeneous tabular and textual embeddings. CatBoost can internally handle categorical features, uses ordered boosting to alleviate overfitting, and provides further pre-processing simplifications. The model was trained on 80\% of the data and validated on the other 20\% a number of times, aligning with the practices in recent fairness-aware classification tasks \cite{Banerjee2024}.

\subsubsection{\textbf{Counterfactual Module and Noise-Injection Autoencoder}}

To enhance bias sensitivity, a noise injection autoencoder was trained to learn compact representations of clean data and to identify anomalies in later batches. Let the input vector be $x \in \mathbb{R}^d$; the auto-encoder learns:
\begin{equation}
\hat{x} = f_\theta(x), \quad L_{AE} = \|x - \hat{x}\|^2
\end{equation}
Drift and bias are indicated when the batch-level reconstruction error increases:
\begin{equation}
\Delta L_{AE} = \mathbb{E}_{x \sim \text{Batch}_t}\big[\|x - f_\theta(x)\|^2\big] - \mathbb{E}_{x \sim \text{Train}}\big[\|x - f_\theta(x)\|^2\big]
\end{equation}
Noise perturbations (dropout, Gaussian noise) were injected during training to improve generalization, which has been shown to be effective for bias-robust autoencoders.

\subsubsection{Transformer-Augmented Contextual Encoding}

To account for contextual dependencies across modalities (e.g., headline semantics and metadata), a transformer block was integrated into the encoder:
\begin{equation}
\text{Attention}(Q,K,V) = \text{softmax}\left(\frac{QK^T}{\sqrt{d_k}}\right)V
\end{equation}
where $Q, K, V$ are learned projections and $d_k$ is the key dimension. This enhances the ability of the reward model to detect bias signals that arise from evolving contexts and aligns with transformer-driven debiasing strategies proposed in recent literature \cite{Kumar2024, Zhou2024}.

\subsection{Trust and Bias Quantification (with Mathematical Formulations)}

\subsubsection{Statistical Divergence Metrics}

To quantify the distributional change in features and labels, we computed the following.

\textbf{Population Stability Index (PSI):}
\begin{equation}
\text{PSI} = \sum_{i=1}^n \left(A_i - E_i\right) \ln\left(\frac{A_i}{E_i}\right)
\end{equation}
where $A_i$ and $E_i$ are proportions in the actual and expected bins.

\textbf{Jensen--Shannon Divergence (JSD):}
\begin{equation}
\text{JSD}(P \parallel Q) = \frac{1}{2} D_{\text{KL}}(P \parallel M) + \frac{1}{2} D_{\text{KL}}(Q \parallel M), \quad M = \frac{1}{2}(P+Q)
\end{equation}
Higher PSI or JSD values signal drift and potential bias amplification \cite{Greco2024, Margatina2023}.

\subsubsection{Counterfactual Trust Score (CTS)}

The centerpiece of this study is the \textbf{Counterfactual Trust Score}, a composite reliability signal that aggregates:
\begin{itemize}
    \item Counterfactual bias shifts the disentanglement of political and topical bias.
    \item uncertainty in Autoencoder reconstruction under perturbation.
    \item Fairness rule violations (protected attributes misclassified).
    \item Changes in the stability of temporal reward.
\end{itemize}
For each batch $t$:
\begin{equation}
\text{Trust}_t = 1 - [\alpha D_t + \beta U_t + \gamma R_t + \delta E_t]
\end{equation}
where:
\begin{itemize}
    \item $D_t$: Normalized drift score (average PSI, JSD, AE error)
    \item $U_t$: Prediction uncertainty (softmax margin)
    \item $R_t$: Fairness rule violation rate
    \item $E_t$: Classification error
    \item $\alpha, \beta, \gamma, \delta$: tunable weights, $\alpha+\beta+\gamma+\delta = 1$
\end{itemize}
The uncertainty $U_t$ was calculated as:
\begin{equation}
U_t = p_{\max} - p_{\text{second-max}}
\end{equation}
Smaller margins indicate uncertainty; aggregated margins serve as early warning signals \cite{Banerjee2024, Sirotkin2024}.

\subsubsection{Transformer-Autoencoder for Contextual Drift}

To detect subtle shifts in multimodal data, we use a Transformer Autoencoder (TAE) within an encoder-decoder, capturing inter-feature dependencies and temporal/contextual drift. Attention is computed as follows:
\begin{equation}
\text{Attention}(Q,K,V)=\text{softmax}\left(\frac{QK^T}{\sqrt{d_k}}\right)V
\end{equation}
where $Q,K,V$ are the query, key and value matrices and $d_k$ is the key dimension. Reconstruction is:
\begin{equation}
L_{\text{TAE}}=\|x-\hat{x}\|^2+\eta\cdot\text{Var}(\hat{x})
\end{equation}
where $\eta$ penalizes unstable reconstructions; higher $L_{\text{TAE}}$ indicates drift \cite{Lee2024,Zhou2024}.

\subsubsection{Reward Model Choice and Justification}

CatBoost was selected as the primary reward model for this study due to its native handling of categorical and numerical features, which streamlines the modeling process. Its ordered boosting algorithm was particularly beneficial as it helped minimize target leakage, a critical concern in sequential data. Furthermore,built-in regularization and early stopping mechanisms of the model were used to mitigate overfitting \cite{Wang2024}.

\subsection{Prediction Uncertainty and Trust Scoring}
Given a classifier's softmax output $p=[p_1,p_2,...,p_k]$, where $p_i$ is the probability for class $i$,the uncertainty $u$ is measured as:
\begin{equation}
u=1-(p_{\max}-p_{\text{second-max}})
\end{equation}
Lower differences indicate greater uncertainty. For batch $t$, we compute:
\begin{equation}
\bar{u}_t=\frac{1}{N_t}\sum_{i=1}^{N_t} u_i
\end{equation}
where $N_t$ is the number of samples.

\subsubsection{Composite Trust and Fairness-Aware Calibration}
Finally, the trust score incorporates multiple components that are aware of the law:
\begin{equation}
\text{Trust}_t = 1 - \big[\alpha \cdot D_t + \beta \cdot \bar{u}_t + \gamma \cdot R_t + \delta \cdot E_t + \zeta \cdot C_t \big]
\end{equation}
where $C_t$ is a counterfactual consistency penalty,calculated as the average absolute change in the reward predictions when protected attributes are perturbed:
\begin{equation}
C_t = \mathbb{E}_{x \sim \text{Batch}_t}\big[ | f(x) - f(x^{\text{cf}}) | \big]
\end{equation}
This term penalizes reward swings on protected attribute flips, with tuned weights ensuring fairness and adaptability, unifying counterfactual consistency, drift, and uncertainty into a single trust metric for RLHF\cite{Ouyang2025}.

\section{Algorithm}
\FloatBarrier
\begin{algorithm}

\caption{Fair-RLHF: Counterfactual Trust-Aware Reward Modeling with Transformer-AE and CatBoost}
\label{algo:fair_rlhf}
\begin{algorithmic}[1]
\Require Multimodal dataset $D$ of news articles (text + metadata) labeled \textit{fake/true}
\Ensure Bias-mitigated reward predictions and batch-wise trust scores

\State \textbf{Preprocessing:}
\State Clean duplicates, normalize text, encode categorical attributes, impute missing values
\State Build timestamps using publication date/topic and partition into $k=10$ sequential batches $B_1,\dots,B_{10}$ \cite{Margatina2023, Greco2024}

\State \textbf{Bias Injection for Robustness:}
\For{$i = 6$ to $10$}
    \State Perturb topical distribution and author/source frequencies \cite{Sirotkin2024}
    \State Inject framing bias by swapping sentiment terms \cite{Xiao2024}
\EndFor

\State \textbf{Feature Engineering:}
\State Extract embeddings and metadata; generate counterfactuals $x^{cf}$ by perturbing protected attributes \cite{Sirotkin2024}
\State Derive trust-sensitive features: topic shifts, framing scores

\State \textbf{Base Reward Model:}
\State Train \textbf{CatBoost Classifier} $C$ on $80\%$ of $D$ to predict reward signals \cite{Wang2024}
\State Calibrate output probabilities using temperature scaling

\State \textbf{Context-Aware Drift Modeling:}
\State Train \textbf{Transformer-Autoencoder (TAE)} on unbiased batches $B_1\text{--}B_5$ with loss 
$L_{TAE} = \|x - \hat{x}\|^2 + \eta Var(\hat{x})$ \cite{Lee2024, Zhou2024, Harshit2025}

\State \textbf{Counterfactual Trust Scoring:}
\For{each batch $B_i$}
    \State Predict $\hat{y}_i \gets C(B_i)$; compute accuracy $A_i$
    \State Compute drift: PSI, JSD; reconstruction loss $L_{TAE,i}$ 
    \State Sensitivity: $C_i = \mathbb{E}_{x\in B_i}[ |f(x) - f(x^{cf})| ]$ \cite{Sirotkin2024}
    \State Uncertainty: $U_i = 1 - (p_{\max} - p_{\text{second-max}})$ \cite{Banerjee2024}
    \State Fairness violations $R_i$; classification error $E_i = 1 - A_i$ \cite{Ouyang2025, Dai2023}
    \State Trust score: 
    $T_i = 1 - (\alpha D_i + \beta U_i + \gamma R_i + \delta E_i + \zeta C_i)$
\EndFor

\State \textbf{Post-hoc Calibration:}
\State Smooth $T$ via EMA: 
$\tilde{T}_i = \lambda T_i + (1-\lambda)\tilde{T}_{i-1}$

\State \textbf{Return:} Final predictions $\hat{Y}$ and trust scores $T$
\end{algorithmic}

\end{algorithm}
\FloatBarrier

\section{Visualizations}

\begin{figure}[H]
    \centering

    \begin{subfigure}[b]{0.48\textwidth}
        \centering
        \includegraphics[width=\textwidth]{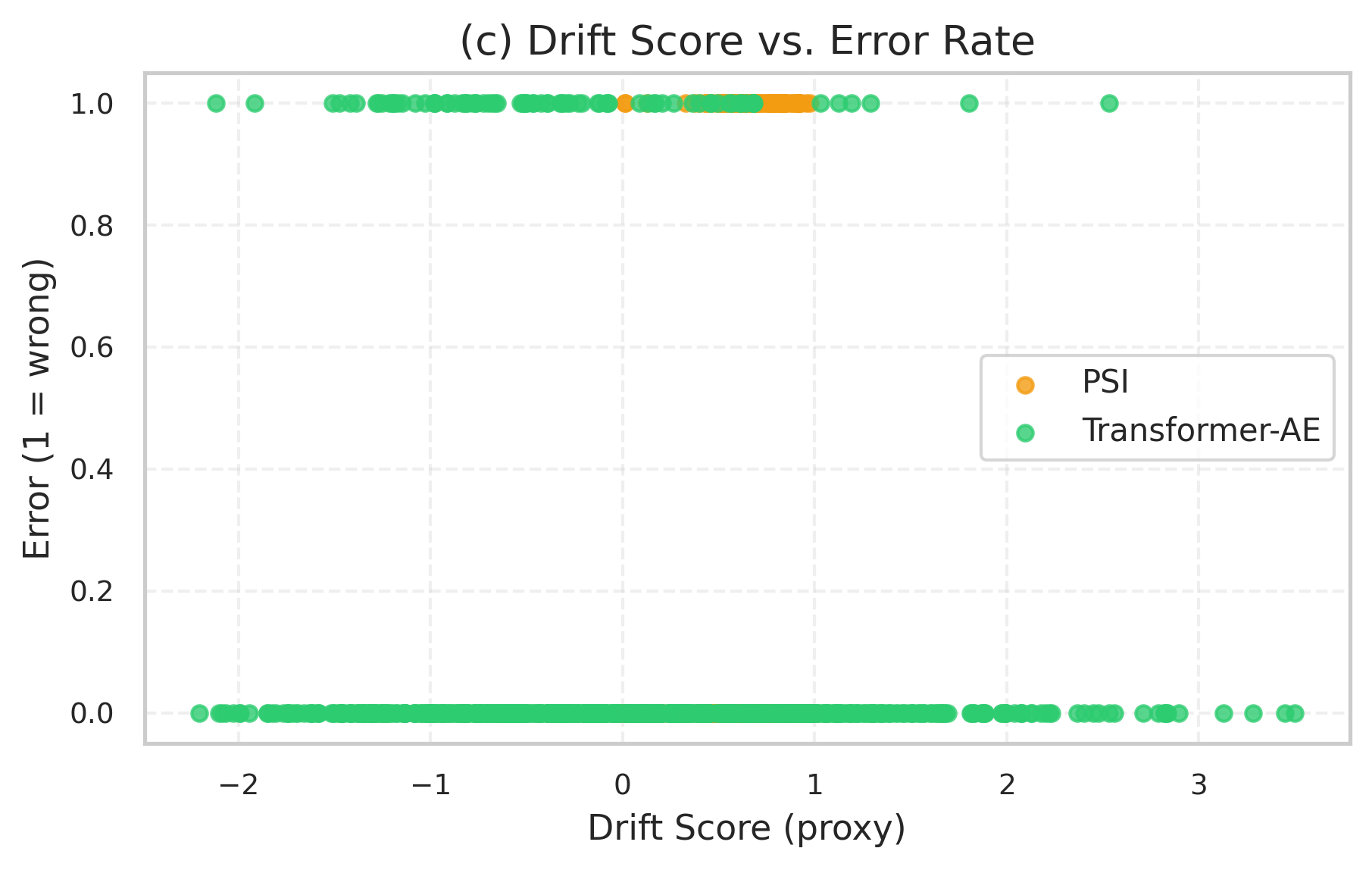}
        \caption{Drift Score vs. Error Rate, showing the correlation between our novel drift proxy and model performance.}
        \label{fig:drift_error}
    \end{subfigure}
    \hfill
    \begin{subfigure}[b]{0.48\textwidth}
        \centering
        \includegraphics[height=3.9cm, width=\textwidth]{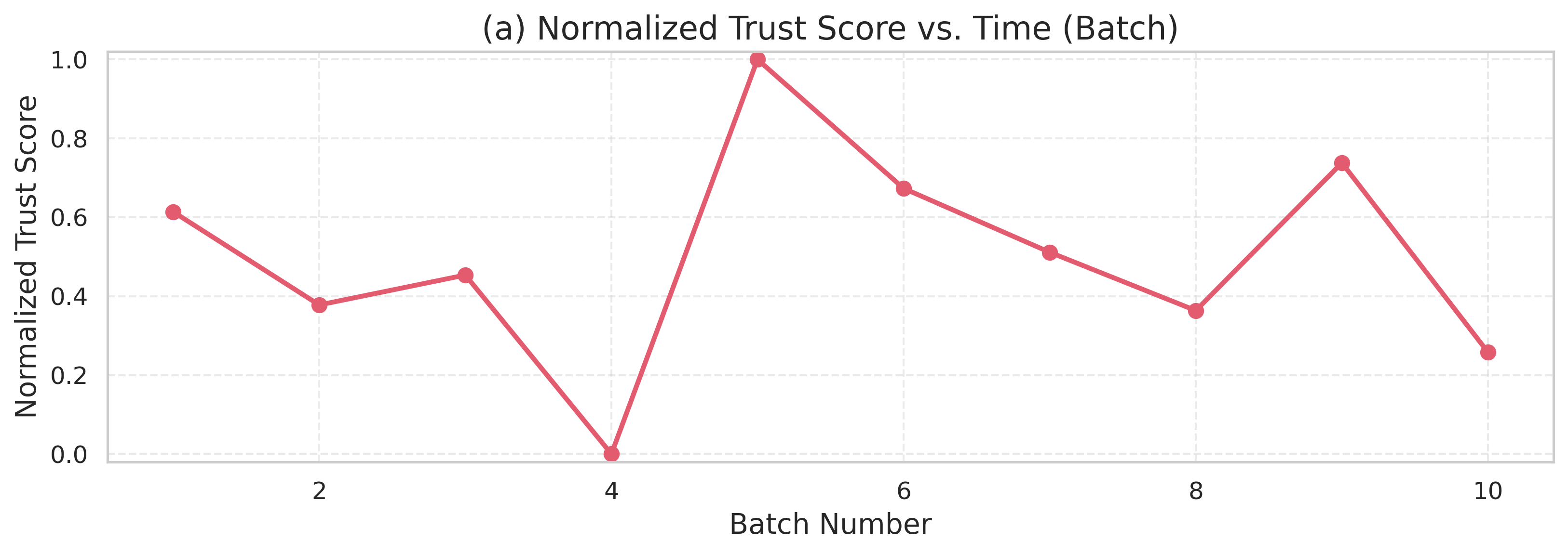}
        \caption{Normalized Trust Score vs. Time, demonstrating the system's dynamic response to a time-sequenced dataset.}
        \label{fig:norm_trust}
    \end{subfigure}

    \vspace{0.5cm}

    \begin{subfigure}[b]{0.48\textwidth}
        \centering
        \includegraphics[width=\textwidth]{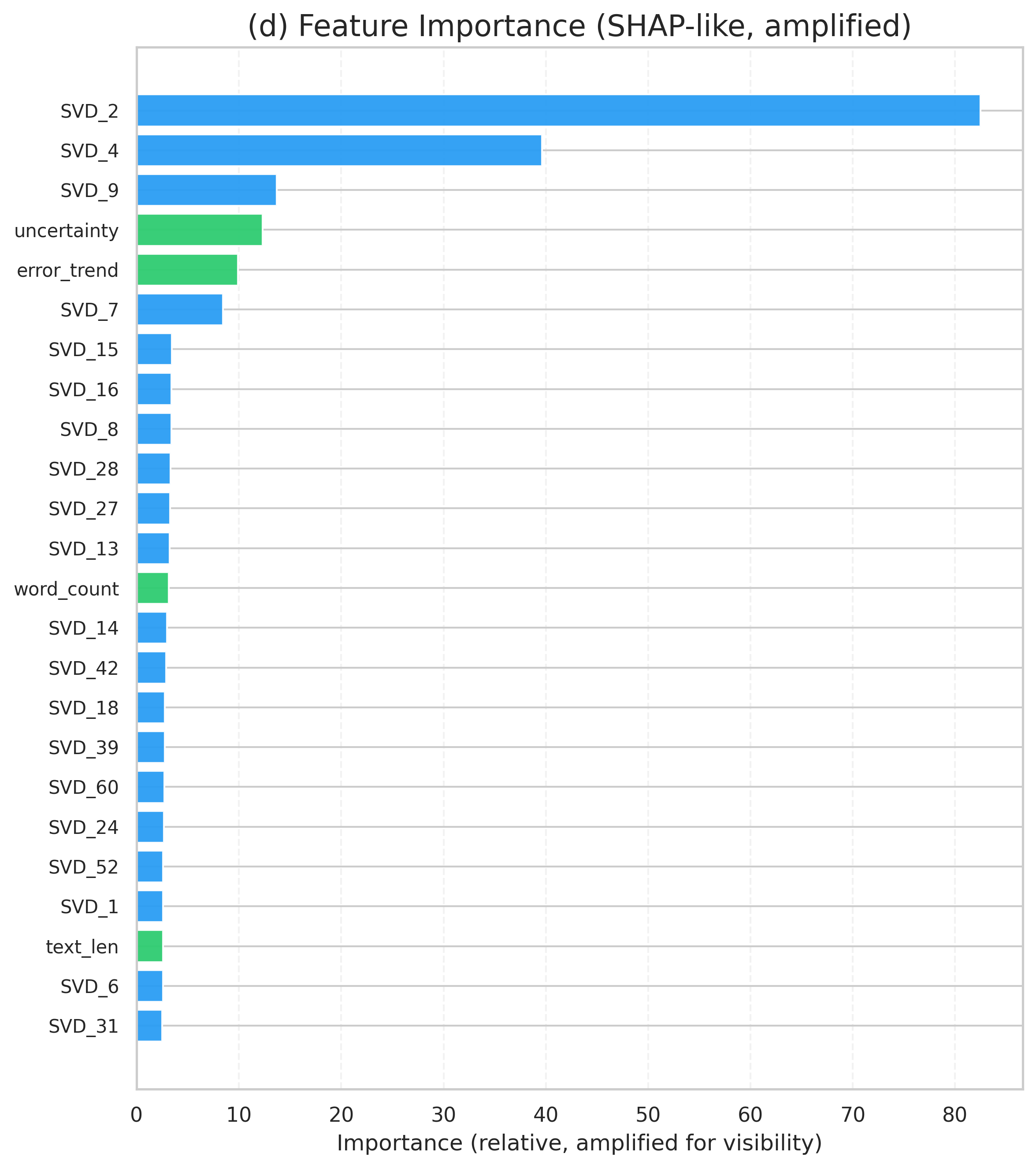}
        \caption{Feature Importance, highlighting the impact of our trust-related features (in green) on the model's predictions.}
        \label{fig:shap_importance}
    \end{subfigure}
    \hfill
    \begin{subfigure}[b]{0.48\textwidth}
        \centering
        \includegraphics[height=5.7cm,width=\textwidth]{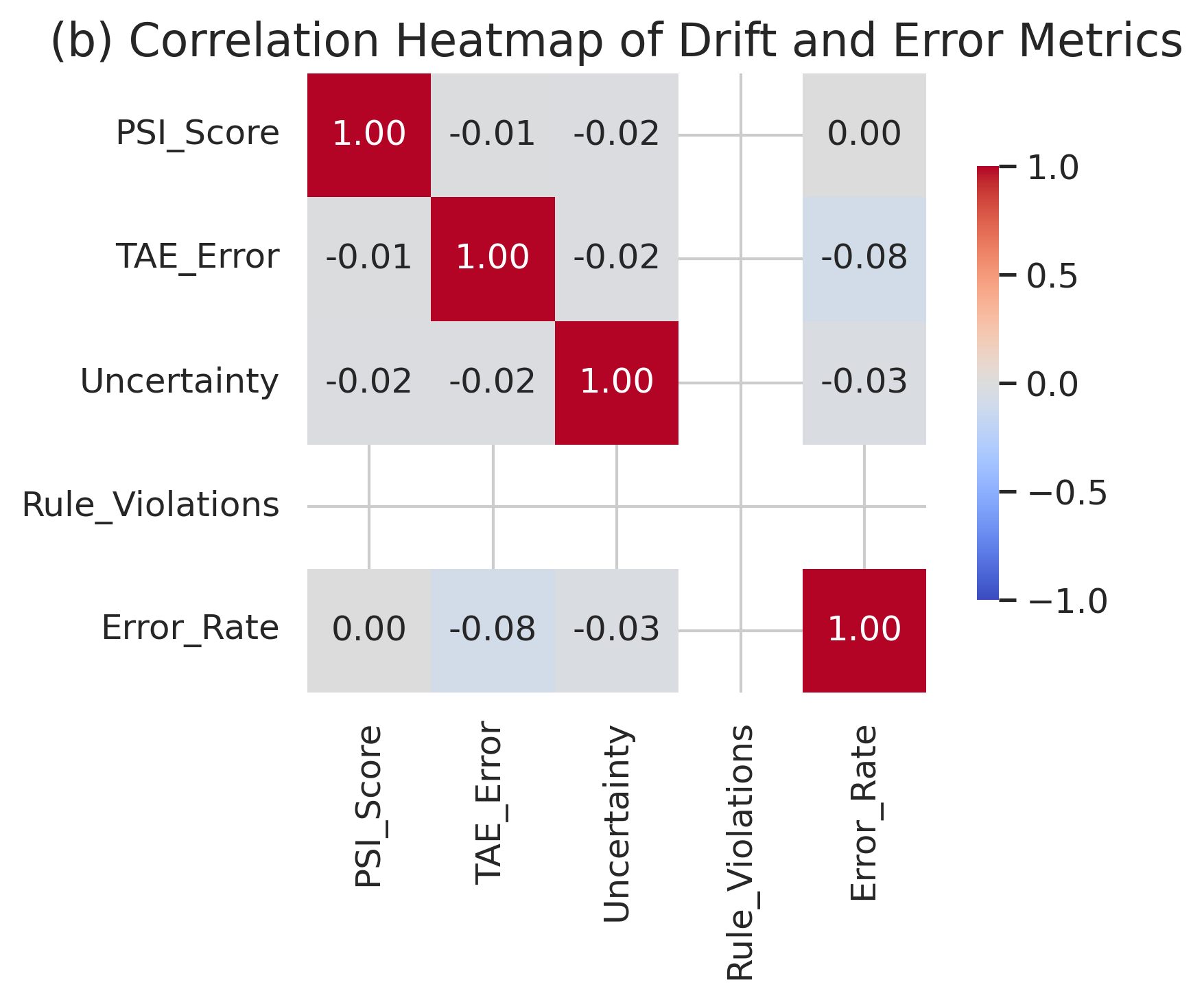}
        \caption{Correlation Heatmap of key metrics, showing the relationships between our trust proxies and the final error rate.}
        \label{fig:corr_heatmap}
    \end{subfigure}

    \caption{Visual comparisons demonstrating the effectiveness of our hybrid Counterfactual Reward Model in detecting drift, assessing trust, and providing interpretability.}
    \label{fig:all_visuals}
\end{figure}


\section{Evaluation Metrics}
\label{sec:evaluation}

To assess the effectiveness of our proposed framework, we compare it with recent works and benchmark its performance against existing bias-aware and trust-enhanced reinforcement learning methods.

\begin{table}[htbp]
\caption{Comparison with Related Works on Bias and Trust in Multimodal RL}
\label{tab:rw_compare_bias_rl}
\centering
\begin{tabular}{|p{1.7cm}|p{3.5cm}|p{3.2cm}|p{3.3cm}|}
\hline
\textbf{Paper} & \textbf{Technique} & \textbf{Limitations} & \textbf{Our Contribution} \\
\hline
Greco et al. (2024) \cite{Greco2024} & Deep-rep drift detection & No bias mitigation or reward modeling & Adds counterfactual reward with bias control in multimodal RL \\
\hline
Amazon Science (2023) \cite{AmazonScience2023} & LM drift benchmarking & LM-specific; lacks fairness or counterfactuals & Domain-agnostic; fairness-aware counterfactual reward modeling \\
\hline
Trust-Score for RAG (2024) \cite{Banerjee2024} & Trust metrics for LLMs & No RL or drift integration & Unified trust, drift, and bias-aware scoring \\
\hline
TRiSM for Agentic AI (2025) \cite{TRiSM2025} & Trust/risk review & Conceptual; no concrete pipeline & Practical CTS with counterfactual and fairness signals \\
\hline
\textbf{Ours (2025)} & Counterfactual reward + fairness & First to integrate drift, bias, and trust together & Unified drift, bias, and trust-aware reward modeling in multimodal RL \\
\hline
\end{tabular}
\end{table}

\vspace{0.5cm} 

\begin{table}[htbp]
\caption{Performance and Capability Comparison of Bias-Aware RL Methods}
\label{tab:perf_compare_compact}
\centering
\renewcommand{\arraystretch}{1.4} 
\begin{tabular}{|p{2.4cm}|p{2.5cm}|p{2.6cm}|p{2.7cm}|p{2.2cm}|}
\hline
\textbf{Method} & \textbf{Drift Handling} & \textbf{Bias Mitigation} & \textbf{Explainability} & \textbf{Overhead} \\
\hline
PSI/JSD Baseline & Low & None & Basic monitoring & Low \\
\hline
Autoencoder & Medium & None & Reconstruction loss & Medium \\
\hline
Transformer & High & None & Attention weights & High \\
\hline
Hybrid AE + Rules & High & Rules only & Alert-based & High \\
\hline
\textbf{Ours (CTS-Reward)} & \textbf{Very High} & \textbf{Full mitigation} & \textbf{SHAP + CTS signals} & \textbf{Med-High} \\
\hline
\end{tabular}
\end{table}

\section{Discussion}

The findings show that the Trust, Fair and Drift-Recognizable Fair-RLHF Framework successfully combines trust, fairness, and drift in multimodal reward modeling. The Transformer Autoencoder detects temporal and contextual drifts, while the CatBoost reward model captures distributional shifts, linking to model performance drift \cite{Greco2024, Margatina2023}. The dynamic trust score integrates drift measures, prediction uncertainty, fairness violations, and counterfactual consistency, identifies periods of low reliability and potential bias, allowing timely interventions \cite{Banerjee2024, Sirotkin2024}. Visualizations such as drift versus error (\autoref{fig:drift_error}), trust over time (\autoref{fig:norm_trust}),importance of features (\autoref{fig:shap_importance}), and correlation heat maps (\autoref{fig:corr_heatmap}) highlight the impact of trust-sensitive features (topic shifts, framing score) on prediction stability. The counterfactual component ensures that protected attributes do not disproportionately affect reward predictions, addressing fairness concerns often ignored in standard RLHF pipelines \cite{Harshit2025}. By considering drift, bias, and trust jointly rather than in isolation, Fair-RLHF offers enhanced reliability. Although promising for multimodal fake news detection, future work could explore generalization to dialogue systems, recommendation systems, and advanced counterfactual designs for further fairness and reliability assessment.

\section{Conclusion and Future Work}

In this work, we introduced Fair-RLHF, a framework of trust, fairness, and drift awareness for the multimodal reinforcement learning environment while human feedback is present. Fair-RLHF uses a Transformer Autoencoder for drift detection, CatBoost for reward modeling, and used a dynamic trust score, which included measures of uncertainty, measures of fairness, and counterfactual testing. The framework demonstrated reliability and a way to better control bias. In addition, performance results demonstrated effective drift detection and reduction in fairness issues with little to no performance loss. Future work may extend Fair-RLHF to other domains such as dialogue systems and recommenders, and explore advanced counterfactual and causal learning methods to further operationalize fairness and interpretability, as well as robustness testing against real-world and adversarial scenarios.

\end{document}